 \documentclass[final,5p,times,twocolumn,authoryear]{elsarticle}

\usepackage{lipsum}
\usepackage{graphicx}
\usepackage{setspace}
\usepackage[subfigure]{tocloft}
\usepackage{siunitx}
\usepackage{floatrow}
\usepackage{longtable}
\usepackage{amssymb,epsfig,amstext,amsmath,subfigure,verbatim}
\usepackage{algorithm}
\usepackage{algpseudocode}

\usepackage{listings}
\usepackage{color}
\usepackage{bbding}
\journal{Computerized Medical Imaging and Graphics}

\begin{document}

\begin{frontmatter}



\title{Single color digital H\&E staining with In-and-Out Net}


\author[inst1]{Mengkun Chen}
\author[inst1]{Yen-Tung Liu}
\author[inst1]{Fadeel Sher Khan}
\author[inst2]{Matthew C. Fox}
\author[inst2]{Jason S. Reichenberg}
\author[inst2]{Fabiana C.P.S. Lopes}
\author[inst2]{Katherine R. Sebastian}
\author[inst1,inst3]{Mia K. Markey}
\author[inst1]{James W. Tunnell\thanks{Corresponding author}}
\affiliation[inst1]{organization={University of Texas at Austin, Department of Biomedical Engineering},%Department and Organization
            addressline={107 W Dean Keeton St}, 
            city={Austin},
            postcode={78712}, 
            state={TX},
            country={United States}}
\affiliation[inst2]{organization={The University of Texas at Austin, Division of Dermatology, Dell Medical School},%Department and Organization
            addressline={1301 Barbara Jordan Blvd \#200}, 
            city={Austin},
            postcode={78732}, 
            state={TX},
            country={United States}}
\affiliation[inst3]{organization={The University of Texas MD Anderson Cancer Center, Department of Imaging Physics},%Department and Organization
            addressline={1400 Pressler Street}, 
            city={Houston},
            postcode={77030}, 
            state={TX},
            country={United States}}

\begin{abstract}
Digital staining streamlines traditional staining procedures by digitally generating stained images from unstained or differently stained images. While conventional staining methods involve time-consuming chemical processes, digital staining offers an efficient and low-infrastructure alternative. Researchers can expedite tissue analysis without physical sectioning by leveraging microscopy-based techniques, such as confocal microscopy. However, interpreting grayscale or pseudo-color microscopic images remains challenging for pathologists and surgeons accustomed to traditional histologically stained images. To fill this gap, various studies explore digitally simulating staining to mimic targeted histological stains. This paper introduces a novel network, In-and-Out Net, designed explicitly for digital staining tasks. Based on Generative Adversarial Networks (GAN), our model efficiently transforms Reflectance Confocal Microscopy (RCM) images into Hematoxylin and Eosin (H\&E) stained images. Using aluminum chloride preprocessing for skin tissue, we enhance nuclei contrast in RCM images. We trained the model with digital H\&E labels featuring two fluorescence channels, eliminating the need for image registration and providing pixel-level ground truth. Our contributions include proposing an optimal training strategy, conducting a comparative analysis demonstrating state-of-the-art performance, validating the model through an ablation study, and collecting perfectly matched input and ground truth images without registration. In-and-Out Net showcases promising results, offering a valuable tool for digital staining tasks and advancing the field of histological image analysis.
\end{abstract}




\end{frontmatter}

%\tableofcontents



\section{Introduction}
\label{introduction}

Digital staining refers to methods that use computational models to generate histological stained images from other images. In histology image analysis, the traditional approach uses dyes to chemically stain frozen sectioned or formalin fixed paraffin embedded tissue block sections like Hematoxylin and Eosin (H\&E) staining. In clinical practice, histological stains such as Hematoxylin and Eosin (H\&E) are vital for the microscopic examination of tissue samples, allowing pathologists to diagnose diseases like cancer, identify infection, and assess tissue abnormalities. However, this fixing-sectioning-staining-diagnosing process is time-consuming and highly infrastructure-demanding. For instance, in some skin surgeries such as Mohs microscopic surgery, cancerous tissue is removed in an iterative process, with each layer removal referred to as a stage. For Mohs surgery, the process from fixation to diagnosis can require at least 30-60 mins for each stage, requiring skilled technicians and substantial equipment.

To eliminate the need for fixation and staining, microscope-based methods(\cite{Longo2014,Patel2007,Abeytunge2013,Yoshitake2018,Glaser2017,Austin2016,Cui2018}) offer a practical substitute for traditional histological staining. By swiftly staining fresh tissue(\cite{Chen2021}) or even eliminating the staining step altogether(\cite{Rivenson2019}), these approaches bypass certain time-consuming and infrastructure-intensive stages. Moreover, they enable bulk tissue imaging without the need for physical sectioning(\cite{Boktor2022}). Consequently, these methods expedite the process and require substantially less infrastructure, usually requiring only a single microscope device(\cite{Boktor2022}).

Despite the efficiency of the microscopic approaches, they have not been widely adopted into clinical practice. This is likely because their visual output looks so different than traditional pathology, that the clinicians can't utilize them without additional training. Pathologists and surgeons are accustomed to interpreting colored histologically- stained images rather than grayscale or pseudo-color microscopic images. Addressing this challenge, numerous studies(\cite{Bai2023}) have explored using computers to digitally simulate the staining of microscopic images to resemble targeted histological stained images. Digital staining offers a practical alternative by allowing for the generation of stained-like images without chemical stains. Furthermore, there is a growing demand for obtaining various types of stained images from a single tissue sample. For instance, researchers may require both H\&E and Masson’s Trichrome (TRI) stained images from the same live biopsy section to examine detailed structures comprehensively(\cite{Foot1933}). This necessity was met by stain-to-stain digital staining, a technique that leverages computers to transform images from one histological stain domain to another(\cite{Bai2023,kevin2021}). The process of converting unstained microscopic grayscale or pseudo-color images to histologically stained images is referred to as "label-free" digital staining. In contrast, labeled digital staining involves converting samples with chemicals into pseudo-color histological images. Additionally, transforming images from one histological stain domain to another is termed "stain-to-stain" digital staining. While these techniques serve different purposes, they share similar underlying approaches.

In conclusion, digital staining aims to produce images that resemble traditional histologically stained slides in a fraction of the time. Techniques like RCM can be used in real-time surgical settings, where quick decision-making is crucial. Digital staining eliminates several steps in the traditional process, reducing time and infrastructure requirements. In these scenarios, a microscope and computational model can replace the complex staining procedures, making digital staining an attractive option for use in operating rooms or point-of-care diagnostics.

Another significant advantage is the ability of digital staining to address non-invasive diagnostic needs. Pathologists and clinicians are accustomed to interpreting histologically stained images, which provide a wealth of diagnostic information. However, in procedures where tissue biopsies are impractical, such as in vivo imaging using techniques like RCM or optical coherence tomography (OCT), the grayscale or pseudo-color images produced by these modalities are often challenging to interpret. Digital staining bridges this gap by converting these grayscale images into colorized images that resemble traditional H\&E-stained tissue. This enhances the diagnostic utility of non-invasive imaging, especially for identifying conditions like basal cell carcinoma or melanoma without the need for a physical biopsy.

Digital staining also has the potential to improve telepathology and remote diagnostics. 
In resource-limited regions, removed skin tissue must be packaged in formalin, shipped to a lab, then processed and stained. Sometimes, this histopathology would then need to be scanned into a digital image and sent to a pathologist at another site. By acquiring digital images in the clinic that can be quickly sent to anyone in the world for interpretation, several steps (and costs) can be eliminated, making diagnostic workflows faster and more accessible. This is especially valuable for remote consultations where physical slide preparation and transport would cause delays.

The initial exploration into digital staining introduced a method that involves the linear combination of two fluorescence confocal microscope channels to produce H\&E-like images.(\cite{Bini, Gareau2009_AO_SR101}) Specifically, one channel mimics the hematoxylin (H) stain, while the other mimics the eosin (E) stain, with each channel capturing light within a specific wavelength range to represent a distinct color in the final image. In this technique, researchers employed two distinct dyes for staining nuclei and collagen/cytoplasm, each corresponding to a separate channel. Each channel was assigned different R, G, and B weights, enabling their amalgamation into an RGB image. Subsequently, Beer's law was applied to nonlinearly combine these channels, enhancing the realism of the resulting H\&E image. Notably, channels can be independent of the fluorescence modality. Alternatively, a reflectance channel can be used to depict collagen/cytoplasm, while another fluorescence channel highlights nuclei.(\cite{Gareau2009,Gareau2012}) As long as one channel captures collagen/cytoplasm and the other showcases nuclei, this methodology remains applicable. This versatile approach sometimes establishes ground truths for digital staining across various modalities. For instance, when digitally staining RCM to H\&E images, Li et al. (\cite{Li2021}) generated H\&E ground truths from two channels, utilizing the approach to circumvent the need for actual chemical H\&E staining.

The rise of artificial intelligence has spurred the development and deployment of numerous generative deep-learning models in the innovative field of digital staining. Among the array of generative models, the prevalent choice for digital staining applications is the utilization of Generative Adversarial Networks (GAN)(\cite{GAN_goodfellow}). Ozcan's laboratory, in particular, has spearheaded groundbreaking research in this domain. Riverson et al.(\cite{Rivenson2019,Rivenson2019_nature}) used pix2pixGAN(\cite{pix2pix2017}) to conduct some remarkable and clinically significant label-free digital staining studies. They digitally stain Quantitative Phase Images (QPI)(\cite{Rivenson2019}) and auto-fluorescence images(\cite{Rivenson2019_nature}) into multiple histological stained counterparts, including H\&E, TRI stain and Jones methenamine silver (JMS) stain. Their tissue types included human skin, gland, kidney, liver, thyroid, and lung. Continuing Riverson’s work, De Hann et al.(\cite{kevin2021}) pioneered the use of CycleGAN(\cite{CycleGAN2017}) in stain-to-stain tasks, enabling the transformation of H\&E-stained images into a diverse array of histological stains, including TRI stain, JMS stain and Periodic acid–Schiff (PAS) stain. Furthermore, Yang et al.(\cite{Yang2022}) combined the techniques of Riverson and De Hann to employ Cascaded GAN networks to stain auto-fluorescence images into PAS-stained images digitally. They first conducted label-free staining using pix2pixGAN to stain auto-fluorescence images into H\&E-stained images; then, they used CycleGan to stain H\&E images into PAS-stained images. They proved this two-step approach performed better than directly stained auto-fluorescence images into PAS-stained images. The works done by Riverson, De Hann, and Yang et al. excellently exemplify the outstanding applications of GAN-based models in the digital staining field. These models can digitally stain diverse human tissue samples and generate various histological stain styles. Beyond GAN-based models, noteworthy strides have been made in other generative models. Lotfollahi et al.(\cite{Lotfollahi2019}) employed a straightforward encoder-decoder architecture, transforming Fourier transform infrared (FTIR) spectroscopic images into H\&E and DAPI (4,6-diamidino-2-phenylindole) images. Burlingame et al.(\cite{Burlingame2020}) used a Variational Autoencoder (VAE)(\cite{kingma2022autoencoding}) network to stain H\&E-stained images into immunofluorescent images. Researchers also tried modifying state-of-the-art models, like GAN-based models, to perform better on digital staining tasks. Li et al.(\cite{Li2021}) integrated attention blocks into the pix2pixGAN model, tailoring them to stain RCM images into the H\&E-stained images. This marked a successful attempt to modify the network for the digital staining tasks. Additionally, Liu et al.(\cite{VSGD}) introduced the VSGD-Net for staining H\&E to Sox10 images, exhibiting outstanding performance in melanocyte detection. These diverse approaches collectively contribute to the multifaceted landscape of generative models, expanding the possibilities in digital staining research.

While state-of-the-art models have garnered notable success in digital staining, only a limited number of models are explicitly designed for digital staining tasks. This paper addresses this gap by introducing a novel network crafted for digital staining. The effectiveness of our proposed model is demonstrated through its successful application in staining RCM images to produce H\&E-Stained images.

This study focuses on employing GAN-based models to achieve grayscale-to-H\&E digital staining. Over the past few decades, diverse microscope-based techniques, including confocal microscopy(\cite{Longo2014,Patel2007,Abeytunge2013}), MUSE(\cite{Yoshitake2018}), light-sheet microscopy(\cite{Glaser2017}), CARS(\cite{Austin2016,Cui2018}), and SRS microscopy(\cite{Austin2016,Cui2018}), have been investigated to speed up the Mohs tissue processing. Among these, RCM stands out as a leading optical imaging method with a commercial device (VivoScope, Caliber I.D.) on the market with US reimbursement codes, making it a forefront contender for surgical guidance development. However, the utilization of RCM for digital staining has been minimally explored due to its limited bright nuclei contrast and its single-channel nature. RCM captures information in a unified manner, making it challenging to separate nuclei and collagen/cytoplasm images. A significant contribution in this domain comes from the work of Li et al.(\cite{Li2021}). They pioneered a two-stage digital staining process for RCM images. Initially, they transformed original RCM images into acetic acid-stained RCM images, enhancing the bright nuclei contrast. Subsequently, they extended this methodology by digitally staining acetic acid-stained RCM images to simulate H\&E images, presenting an innovative approach within the field.

This study introduces a GAN-based architecture to transform single-channel RCM images into H\&E-stained images. To enhance nuclei contrast under RCM, we employed aluminum chloride as a preprocessing agent for skin tissues. To train the model, we utilized digital H\&E labels with two fluorescence channels to provide pixel-level ground truth without needing to perform image registration. Our contributions can be succinctly summarized as follows:
\begin{enumerate}
    \item \textbf{Introduction of In-and-Out Net:} We proposed the In-and-Out Net, a model meticulously designed for label-free digital staining tasks where the training ground truth H\&E images are synthesized through two channels. It also applies when the ground truth images are chemically stained and can be decomposed into two channels representing different dyes. The model is structured with an inner loop and an outer loop, each serving distinct roles in the digital staining process. Specifically:
    \begin{itemize}
        \item \textbf{Inner Loop (Detailed Morphological Big-Tuning):} The inner loop is responsible for the \textbf{detailed morphological big-tuning} of the image. This loop focuses on accurately capturing large-scale morphological features, such as cell boundaries, nuclei, and overall tissue structure. It operates primarily on the structural aspects of the image, ensuring that the major morphological elements are correctly represented before moving on to finer adjustments. By splitting the image into separate H and E channels, the inner loop allows the model to isolate and refine important structures better, ensuring that these morphological details are clear and precise.
        
        \item \textbf{Outer Loop (Fine-Tuning of Morphology and Color):} The outer loop performs \textbf{fine-tuning}, not only on the morphological details but also on the \textbf{color properties} of the image. Once the inner loop has established a strong foundation of structural information, the outer loop adjusts the finer details of the morphology, such as smoothing out boundaries, refining textures, and ensuring small-scale accuracy. In addition, the outer loop handles the accurate \textbf{colorization} of the image, combining the H and E channels to create a final output with the realistic colors and contrast expected from H\&E staining. This step is crucial in ensuring that the color harmonizes well with the morphology, as accurate color is essential for diagnostic purposes in histology.
    \end{itemize}
    
    \item \textbf{Tailored for digital H\&E Staining:} Unlike traditional image-to-image translation models, our model is specifically designed to address the unique challenges of digital staining. The in-and-out loop framework ensures that morphological and color properties are learned separately but integrated smoothly for better accuracy in the final output.

    \item \textbf{Comprehensive Comparative Analysis and Ablation Study:} We comprehensively compared our model and previous approaches for label-free and stain-to-stain digital staining. The results demonstrate that our model achieves state-of-the-art performance. Additionally, we performed an extensive ablation study to validate the contribution of each component in the architecture. The study shows that combining the inner loop, outer loop, and H\&E channel separation leads to \textit{significantly improved image quality}, both in structure and color.

    \item \textbf{Data Collection and Model Validation:} We collected reflectance channel images and two channels of fluorescence images using the same microscope, creating a pixel-to-pixel match between input and ground truth images without the need for registration. This precise data collection, combined with the tailored architecture of our model, ensures the production of highly accurate digital staining results.

    \item \textbf{Optimized for Hierarchical Learning:} The model is optimized to handle both the \textit{structure-color interaction} and the complexity of histological images, ensuring that even complex tissue morphologies are accurately stained digitally. This method addresses the specific needs of \textit{pathology} and \textit{histological image analysis}, maintaining high structural fidelity and realistic color mapping.
\end{enumerate}

\section{Methods and experiments}
\label{sect:sections}
In this section, we describe our data collection, data preprocessing, the design of our proposed In-and-Out Net, and the training approach. In the Results and discussion section, we compare our model results with several state-of-the-art models and ablate In-and-Out Net’s components to validate its design. We also compare the generator loss within different training approaches.

\subsection{Data collection}
\begin{figure*}[tbp]
  \centering
  \includegraphics[width=1\textwidth]{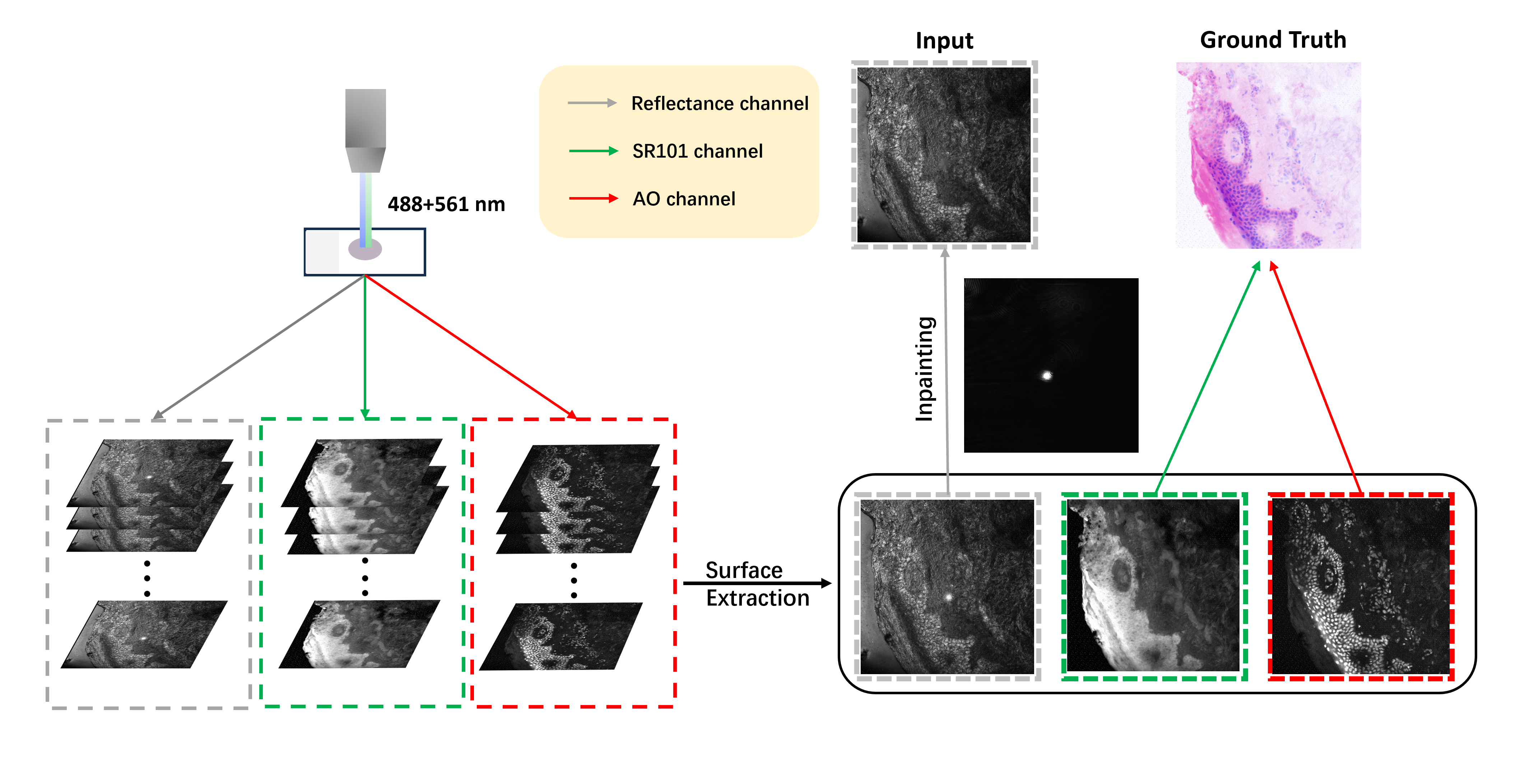}
  \caption{Data collection and preprocessing. The images were collected with three channels simultaneously: reflectance channel (\textless495nm, grey box), SR101 channel (605±70nm, green box) and AO channel (525±50nm, red box). The stack images were conducted with surface extraction to form a single layer image. Reflectance channel images went through a inpainting step to remove the optical interference to generate the input images. Ground truth images were generated by AO channel and SR101 channel images.}
  \label{fig:data collection}
\end{figure*}

Twenty skin tissue samples were collected from sixteen patients undergoing Mohs micrographic surgery. Eight of these samples exhibited BCC and normal tissue, while the remaining twelve contained solely normal tissue. Each sample underwent frozen sectioning of 8 µm thickness at -22℃ with a microtome. After sectioning, a solution composed of acridine orange (AO) and Sulforhodamine 101 (SR101) (8mg AO and 5mg SR101 in 50ml 1:1 water:ethanol solution) was applied for 2 minutes on the sectioned skin samples, followed by 30-second rinse with a 1:1 water:ethanol solution. Afterward, a 30\% aluminum chloride water solution was applied for 1 minute, followed by a rinse with pure ethanol. Aluminum chloride increases nucles contrast under RCM and is a chemical applied to stop bleeding during surgery.(\cite{Scope2010,Flores2015} Following the staining process, confocal fluorescence microscopy mosaic images were obtained using Bruker Prairie View Ultima IV. Each scan covered a 450*450 µm field of view with 512*512 pixels. As illustrated in Figure~\ref{fig:data collection}, each scanning procedure captured 5-15 different depth layers with a step size of 1 µm, resulting in an image stack. The excitation wavelengths were 488nm and 561nm. 488nm laser was used for reflectance mode and AO excitation, and 561nm was used for SR101 excitation. Three channels were employed for data collection: the reflectance channel for wavelengths \textless495nm, the AO channel for wavelengths 525±50nm, and the SR101 channel for wavelengths 605±70nm. The dataset comprises 705 image stacks within each channel. This research was approved by the Institutional Review Board at The University of Texas at Austin and the Seton Healthcare Family. 

\subsection{Data preprocessing}
The image stacks represent different layers of the sample. In many scenarios, achieving a perfectly flat sample surface is challenging, making it difficult to ensure precise alignment of the laser tomography sectioning plane with the sample surface. Consequently, extracting surface information from image stacks becomes necessary. One effective strategy involves leveraging the stacked images as model inputs, allowing the network to discern surface details (\cite{Lotfollahi2019,Li2021}). However, in the future where digital staining can be applied in vivo, patient wounds are unlikely to exhibit the same flatness as sectioned tissue, potentially resulting in each layer containing diminished information compared to sectioned tissue samples. Moreover, using image stacks as model inputs mandates consistency in the number of layers (the number of images within each stack), imposing limitations on the layer numbers. To address this challenge, we adopted the surface extraction algorithm proposed by Gao et al. (\cite{Gao2023}) to extract a single-layer image from each stack. This algorithm calculates the Fourier transformation along depth axis to find the rapid change pixels and determine the surface.

Furthermore, an optical interference issue caused a bright dot to appear in all reflectance channel images (Figure~\ref{fig:data collection}). The Bruker Prairie View Ultima IV system isn't originally designed for reflectance mode. When we replace the dichroic with a beam splitter to enable both fluorescence and reflectance modes, the unintended artifact of reflected light within the optical path emerges within the reflectance channels. To address this, we acquired an additional image specifically capturing the interference dot, serving as a mask. Subsequently, we employed image inpainting techniques to fill in and correct the bright dot in the images. AO and SR101 channel images were transformed into H\&E images as ground truth using the method proposed by Giacomelli et al. (\cite{Giacomelli2016}). During the training phase of our experiment, each image underwent random cropping, resulting in the creation of 10 distinct batches of 256x256 pixels. We collected 705 images from 15 patients (8,225,15,32,105,12,11,103,40,5,13,9,45,12,70 images from each patient, respectively). The variation in numerical data among patients came from the inconsistent imaging quality. As previously mentioned, our imaging system wasn't initially designed for reflectance mode, resulting in significant variability in RCM image quality. We filtered out images based on four criteria: 1. Overexposed and underexposed images with substantial loss of information; 2. Excessive exposure disparities between RCM and fluorescence images; 3. Images lacking clear nuclei contrast under RCM due to poor aluminum chloride staining; 4. Outside region of interest images and corrupted images. Finally, we used 13 patients with 587 images as training set and 2 patients with 118 images as test set. After random cropping, we end up with 5870 images in training set and 1180 images in test set.

The image stacks were generated using ImageJ, while surface extraction, inpainting, and random cropping were implemented using the Python programming language. Both ImageJ and Python codes were executed on an AMD Ryzen 9 5900X CPU.

\subsection{Model Architecture}
The In-and-Out Net architecture is illustrated in Figure~\ref{fig:architecture}. Figure~\ref{fig:architecture}a depicts the overall architecture of the network, while Figure~\ref{fig:architecture}b describes the inner loop and outer loop of the network. Additionally, Figure~\ref{fig:architecture}c provides a detailed structure for both the generator and discriminator.

\subsubsection{Overall architecture}
\begin{figure*}[tbp]
  \centering
  \includegraphics[width=1\textwidth]{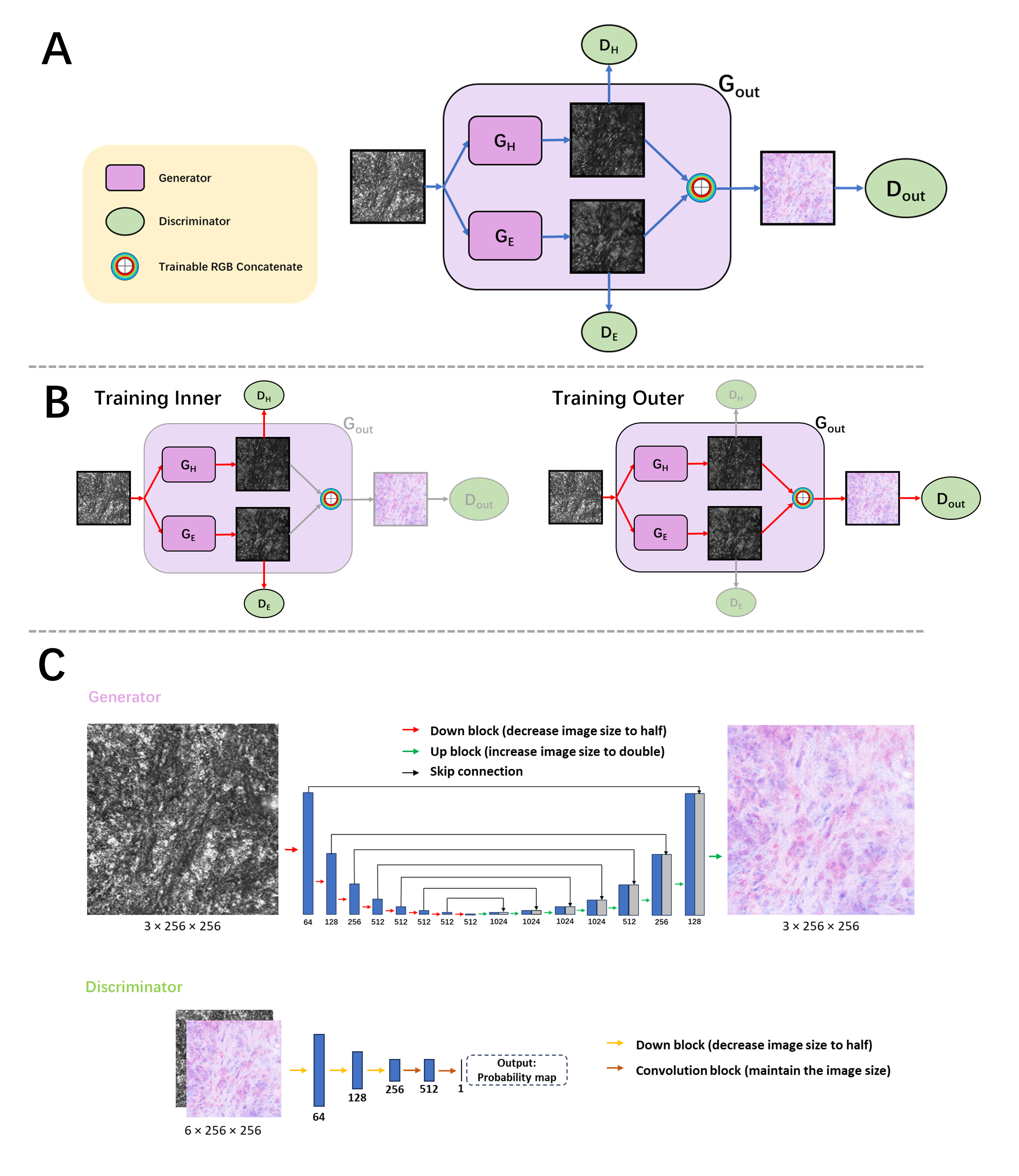}
  \caption{Model Architecture. a) Overall architecture. $G_H$, $G_E$ are generators, $D_H$, $D_E$ and $D_{out}$ are discriminators. $G_H$, $G_E$ and RGB concatenate layer forms $D_{out}$. b) Inner loop and outer loop. The red lines/arrows indicate active for training, while grey lines/arrows indicate inactive for training. c) Architecture for generators and discriminators.}
  \label{fig:architecture}
\end{figure*}

Figure~\ref{fig:architecture}a illustrates the overall architecture of the In-and-Out Net. The network comprises two generators, three discriminators, and a trainable RGB concatenate layer. When dealing with H\&E-stained skin tissue, hematoxylin predominantly stains nuclei in blue-purple, while eosin stains cytoplasm, connective tissue, and extracellular matrix components in pink-red. Our network aims to generate images stained separately with these two chemicals, referred to as the "H channel" image and the "E channel" image for hematoxylin and eosin staining, respectively. The two generators, $G_H$ and $G_E$, are responsible for generating the H and E channel images, each with a corresponding discriminator, $D_H$ and $D_E$. The RGB image is then generated from the H and E channel images using a trainable RGB concatenate layer, as shown in Equation~\ref{eq:rgb layer}.

\begin{align}
  \mathbf{I}_{RGB}(i,j,c) &= 1 - \mathbf{I}_H(i,j) \cdot \text{sigmoid}(W_{H,c} - \mathbf{I}_E(i,j) \cdot \text{sigmoid}(W_{E,c} \label{eq:rgb layer} \\
  \text{sigmoid}(x) &= \frac{1}{1 + e^{-x}} \label{eq: sigmoid}
\end{align}

Here, $\mathbf{I}_{H}(i,j)$ and $\mathbf{I}_{E}(i,j)$ represent the pixel values of the output single-channel images from $G_H$ and $G_E$, respectively. $\mathbf{I}_{RGB}(i,j,c)$ is the pixel value of the output three-channel image from the RGB Concatenate Layer. $W_{H,c}$ and $W_{E,c}$ are trainable variables, where $c=0,1,2$ corresponds to $R,G,B$ channels, respectively. For the output image $\mathbf{I}_{RGB}(i,j,c)$, an additional discriminator $D_{out}$ is employed for supervision. Equation~\ref{eq:rgb layer} can also be written as:
\begin{equation}
    \mathbf{I}_{RGB} = f_{RGB\_Concat}(\mathbf{I}_H, \mathbf{I}_E) = 1 - \mathbf{I}_H \cdot \text{sigmoid}(\mathbf{W}_H) - \mathbf{I}_E \cdot \text{sigmoid}(\mathbf{W}_E) \label{eq: f_RGB}
\end{equation}
Where $\mathbf{W}_H$ and $\mathbf{W}_E$ are both trainable \(3 \times 1\) matrix.

\subsubsection{Inner loop and outer loop}
As depicted in Figure~\ref{fig:architecture}a, $G_H$, $G_E$, $\mathbf{I}_H$, $\mathbf{I}_E$, and the RGB concatenate layer are enclosed within a square box, forming a larger generator denoted as $G_{out}$. Within $G_{out}$, referred to as the inner loop, the combination of $G_H$, $G_E$, $\mathbf{I}_H$, $\mathbf{I}_E$, and the RGB concatenate layer collectively functions as a single entity corresponding to $D_{out}$ in the outer loop.

During the training process, the inner loop and outer loop are trained separately. Figure~\ref{fig:architecture}b illustrates the training of the inner loop, where the training process exclusively follows the red arrows. In this phase, $G_H$ and $G_E$ are trained with the assistance of $D_H$ and $D_E$ to generate $\mathbf{I}_H$ and $\mathbf{I}_E$. The inner loop emphasizes the training of structural information in the output images, excluding considerations of color information and the process of combining the structural images. Consequently, it serves as a more detailed structure-oriented loop.

Conversely, during the training of the outer loop, $G_H$, $G_E$, $\mathbf{I}_H$, $\mathbf{I}_E$, and the RGB concatenate layer collectively form a single generator $G_{out}$. In this loop, the final RGB output image $\mathbf{I}_{RGB}$ is generated with the assistance of $D_{out}$. The outer loop is more concerned with the combination of $\mathbf{I}_H$ and $\mathbf{I}_E$ as well as the color information. Hence, it serves as a loop focusing on overall structures.

\subsubsection{Generator and discriminator architectures}
The generator and discriminator architectures are shown in  Figure~\ref{fig:architecture}c. The generators, $G_H$ and $G_E$, share identical architectures. Both generators employ a U-Net structure based on the pix2pixGAN Generator framework. The U-Net architecture is known for its effectiveness in image-to-image translation tasks, allowing for the preservation of high-level structural features while generating realistic outputs. This design choice enhances the network's ability to capture and reproduce intricate details in the H and E channel images.

On the other hand, the discriminators, $D_H$, $D_E$, and $D_{out}$, also have identical structures. Each discriminator is designed as a straightforward five-layer fully convolution network with a probability map as its output. The discriminators play a crucial role in providing feedback to the respective generators, guiding them towards producing more realistic and accurate outputs.

The shared architectures among generators and discriminators foster consistency and streamline the training process. This uniformity allows for a more cohesive learning experience, ensuring that the network learns general features effectively across the H and E channel generators.

The layer settings, parameter numbers and training settings are present in Table~\ref{tab:layer settings}.

\begin{table*}[ht]
\centering
\caption{Layer settings, activation functions, and optimization methods for In-and-Out Net (Generator and Discriminator)}
\begin{tabular}{|c|c|c|c|c|}
\hline
\textbf{Layer Type (Generator)}         & \textbf{Output Shape} & \textbf{Kernel Size} & \textbf{Stride} & \textbf{Activation Function} \\
\hline
Input Layer                             & (3, 256, 256)         & -                    & -               & -                            \\
\hline
Initial Convolution (Downsample)        & (64, 128, 128)        & 4x4                  & 2               & LeakyReLU                    \\
\hline
Convolutional Layer 1 (Downsampling)    & (128, 64, 64)         & 4x4                  & 2               & LeakyReLU                    \\
\hline
Convolutional Layer 2 (Downsampling)    & (256, 32, 32)         & 4x4                  & 2               & LeakyReLU                    \\
\hline
Convolutional Layer 3 (Downsampling)    & (512, 16, 16)         & 4x4                  & 2               & LeakyReLU                    \\
\hline
Bottleneck                              & (512, 8, 8)           & 4x4                  & 2               & ReLU                         \\
\hline
Transpose Conv Layer 1 (Upsampling)     & (512, 16, 16)         & 4x4                  & 2               & ReLU (with Dropout)          \\
\hline
Transpose Conv Layer 2 (Upsampling)     & (512, 32, 32)         & 4x4                  & 2               & ReLU (with Dropout)          \\
\hline
Transpose Conv Layer 3 (Upsampling)     & (256, 64, 64)         & 4x4                  & 2               & ReLU                         \\
\hline
Final Transpose Conv Layer              & (3, 256, 256)         & 4x4                  & 2               & Tanh                         \\
\hline
\textbf{Total Parameters (Generator)}   & \multicolumn{4}{c|}{108,828,940 (In-and-Out Net)} \\
\hline
\end{tabular}

\vspace{0.5cm}

\begin{tabular}{|c|c|c|c|c|}
\hline
\textbf{Layer Type (Discriminator)}     & \textbf{Output Shape} & \textbf{Kernel Size} & \textbf{Stride} & \textbf{Activation Function} \\
\hline
Input Layer                             & (3, 256, 256)         & -                    & -               & -                            \\
\hline
Convolutional Layer 1                   & (64, 128, 128)        & 4x4                  & 2               & LeakyReLU                    \\
\hline
Convolutional Layer 2                   & (128, 64, 64)         & 4x4                  & 2               & LeakyReLU                    \\
\hline
Convolutional Layer 3                   & (256, 32, 32)         & 4x4                  & 2               & LeakyReLU                    \\
\hline
Convolutional Layer 4                   & (512, 16, 16)         & 4x4                  & 2               & LeakyReLU                    \\
\hline
Convolutional Layer 5                   & (512, 8, 8)           & 4x4                  & 2               & LeakyReLU                    \\
\hline
Final Convolutional Layer               & (1, 1, 1)             & 4x4                  & 1               & Sigmoid                      \\
\hline
\textbf{Total Parameters (Discriminator)} & \multicolumn{4}{c|}{2,765,121} \\
\hline
\end{tabular}

\vspace{0.5cm}

\begin{tabular}{|c|c|}
\hline
\textbf{Optimization Method}  & \textbf{Setting}  \\
\hline
Optimizer                    & Adam \\
Learning Rate                & 0.0002 \\
Beta1 (Momentum term)        & 0.5 \\
Weight Decay                 & 0 (None) \\
Loss Function                & GAN Loss (pix2pix framework) \\
Epochs                       & 400 \\
Batch Size                   & 16 \\
\hline
\end{tabular}
\label{tab:layer settings}
\end{table*}

\subsection{Training and implementation}

\subsubsection{pix2pixGAN Loss}
The optimization of the generator ($G$) and discriminator ($D$) is guided by the pix2pixGAN loss function(\cite{pix2pix2017}):

\begin{align}
  \begin{split}
    L_{\text{pix2pix}}(G, D) = &\mathbb{E}_{x, y}[\log D(x, y)] + \\
                        &\mathbb{E}_{x, z}[\log(1 - D(x, G(x, z)))] + \\
                        &\lambda_0 \cdot \mathbb{E}_{x, y, z}[\|y - G(x, z)\|_1]
  \end{split}
\end{align}

Here, $x$ represents the condition (RCM image), $z$ is the input image (RCM image), and $y$ is the ground truth image (H\&E-stained image). In this context, when $x=z$, the generator takes a single RCM image as input. The generator aims to minimize the overall loss, while the discriminator seeks to maximize it. The first two terms originate from the standard conditional GAN (cGAN)(\cite{conditionalGAN}) loss, and the third term introduces the $\mathcal{L}_{1}$ loss to encourage the generated image to closely resemble the ground truth image. The weight of the $\mathcal{L}_{1}$ loss is denoted as $\lambda_0$.

\subsubsection{Inner and outer loop loss}
During inner loop training, $G_H$, $G_E$, $D_H$, and $D_E$ are optimized by combining two pix2pixGAN loss functions:

\begin{equation}
    L_{\text{in}} = L_{\text{pix2pix}}(G_H, D_H) + L_{\text{pix2pix}}(G_E, D_E)
\end{equation}

In the outer loop, $G_H$, $G_E$, and the RGB concatenate layer are treated as a single generator, denoted as $G_{\text{out}}$. The loss function is expressed as:

\begin{align}
    L_{\text{out}} &= \begin{aligned}[t]
        &L_{\text{pix2pix}}(G_{\text{out}}, D_{\text{out}} + \\
        &\lambda_1 \cdot \mathbb{E}_{x, y}[\|y - G_H(x)\|_1] + \lambda_2 \cdot \mathbb{E}_{x, y}[\|y - G_E(x)\|_1] + \\
        &\mathbb{E}_{x}[\log(1 - D_H(x, G_H(x)))] + \mathbb{E}_{x}[\log(1 - D_E(x, G_E(x)))]
    \end{aligned} \label{eq: L_out} \\
    G_{\text{out}} &= f_{\text{RGB\_Concat}}(G_H, G_E)
\end{align}

The first line in Equation~\ref{eq: L_out} represents the standard pix2pixGAN loss. Additionally, $\mathcal{L}_{1}$ loss terms for both $G_H$ and $G_E$ are included in the second line to preserve structural information during the outer loop. Moreover, $D_H$ and $D_E$ contribute to the third line, serving a role similar to the $\mathcal{L}_{1}$ loss of $G_H$ and $G_E$, preventing significant changes in structural information. However, $D_H$ and $D_E$ are not trained during the outer loop with their weights frozen. The function $f_{\text{RGB\_Concat}}$ is defined in Equation~\ref{eq: f_RGB}.

\subsubsection{Overall loss}
The overall loss is obtained by combining the inner and outer loop losses:

\begin{equation}
    L_{\text{In-and-Out}} = \alpha \cdot L_{\text{in}} + (1 - \alpha) \cdot L_{\text{out}} \label{eq: loss_overall}
\end{equation}
\[
\alpha =
\begin{cases}
    1 & \text{when training the inner loop}, \\
    0 & \text{when training the outer loop}.
\end{cases}
\]
where $\alpha$ is determined based on whether the model is training in the inner loop or outer loop.

Consequently, the final objective is defined as:

\newcommand{\argmax}{\operatornamewithlimits{argmax}}
\newcommand{\argmin}{\operatornamewithlimits{argmin}}
\begin{equation}
    G_{\text{out}}^{*} = \argmin_{G_H, G_E, G_{out}}\argmax_{D_H, D_E, D_{\text{out}}}L_{\text{In-and-Out}}
\end{equation}

\subsubsection{Implementation}
The training process follows a specific pattern, consisting of 10 epochs in the inner loop, followed by 10 epochs in the outer loop, and then another 10 epochs in the inner loop. This alternating training scheme continues until a total of 400 epochs are completed. Throughout the training, the step size remains consistently at $10^{-4}$, and the batch size remains fixed at 16 batches. The Adam optimizer is employed as the trainer for both the inner and outer loops. The code is written in the Python programming language with PyTorch and executed on an Nvidia RTX 3060 GPU for 16 hours.

\subsection{Model evaluation}
\subsubsection{Model comparison}
We compare our In-and-Out Net with four existing models: pix2pixGAN(\cite{pix2pix2017}), pix2pixHD(\cite{pix2pixHD}), VSGD-Net(\cite{VSGD}), and RestainNet(\cite{restainnet}). pix2pixGAN serves as the foundational structure for our network, while pix2pixHD builds upon pix2pixGAN by incorporating additional blocks and introducing object segmentation. VSGD-Net is specifically designed for transforming H\&E-stained images into Sox10 stained images, focusing on the specification of melanocytes. In our adaptation, we replace melanocytes with RCM nuclei to enhance nuclear staining. RestainNet addresses the stain normalization task, transforming one H\&E-stained image into another with a different color style. Its architecture closely resembles ours. Instead of initially training the input image to H and E channel images, RestainNet first trains the input image to an H\&E-stained image. Subsequently, it breaks down this H\&E-stained image into H and E channels, feeding these two channel images back into the loss function. To assess and compare the performance of these models, we utilize metrics such as Mean Squared Error (MSE), Peak Signal-to-Noise Ratio (PSNR), Structural Similarity Index (SSIM), Feature Similarity Index (FSIM), and Multi-Scale Structural Similarity Index (MS-SSIM).

\subsubsection{Ablation study}
The In-and-Out Net model can be dissected into four distinct components for ablation study: 1) Inclusion or exclusion of In-and-Out training, 2) Presence or absence of $D_{out}$, 3) Presence or absence of $D_H$\&$D_E$, and 4) Integration or exclusion of H/E channel branches. In-and-Out training involves the alternating training approach between the inner and outer loops, and H/E branches denote the initial training of RCM into two channels.

\subsubsection{Training approach comparison}
The objective of alternating training between the inner and outer loops is to stabilize structural details (inner loop) and subsequently make fine adjustments (outer loop) to color and channel combinations. To delve deeper into this training method, including its sensitivity to epoch numbers, we conducted a detailed study. To effectively compare the training approaches, monitoring the loss changes during training is essential. However, utilizing the overall loss function from Equation~\ref{eq: loss_overall} alone is inaccurate, given that inner and outer loops are dominated by different loss functions. To address this, we used the following function, denoted as $eval\_loss$:
\begin{equation}
eval\_loss = \mathbb{E}[\|y - G_{out}(x)\|_1] \label{eq: eval_loss}
\end{equation}
Here, $x$ represents the RCM image, and $y$ is the H\&E ground truth image. The $eval\_loss$ function calculates the $\mathcal{L}_{1}$ loss between the output image and the ground truth image, providing a measure of similarity. By tracking $eval\_loss$, we gain insights into how different training approaches impact the model.

\section{Results and discussion}
\subsection{Model results comparison}
\begin{figure*}[th]
  \centering
  \includegraphics[width=1\textwidth]{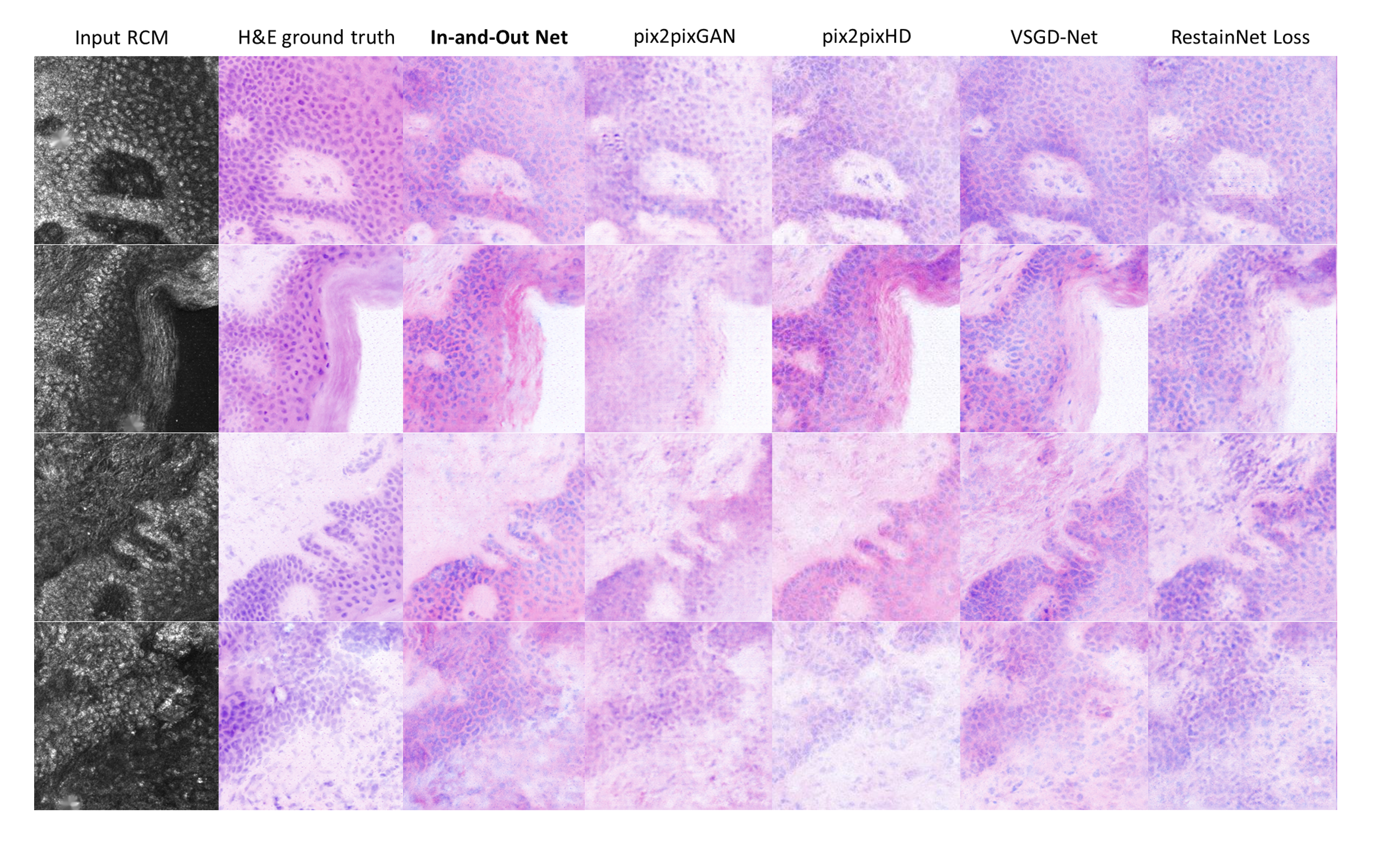}
  \caption{Examples of output images from different models. The generated images in columns from left to right are: Input RCM images, H\&E ground truth, In-and-Out Net, pix2pixGAN, pix2pixHD, VSGD-Net and RestainNet loss. The scale bar is 50 \si{\micro\metre}.}
  \label{fig:results}
\end{figure*}

In Figure~\ref{fig:results}, we can observe example outputs from the different models tested. 

Images generated by pix2pixGAN effectively colorized the input RCM images, although they lacked detail in morphological structures. pix2pixHD successfully captured morphological details and applied appropriate colors, yet the color rendition appears artificial and oversaturated, especially affecting the clarity of nuclei. VSGD-Net combines the strengths of the previous two models but introduces some artifacts in the dermis area. The model is too sensitive to the high valued pixels in RCM images, so it stains most of the collagen structures in dermis area as saturated as dermal/epidermal junction, making the images look dirty. RestainNet loss yields good nucleus appearance, but other structures such as stratum corneum and hair follicles are not well recognized. In comparison, In-and-Out Net provides images with good color representation for all structures and accurately recognizes detailed structures like hair follicles and stratum corneum. It also presents a clean background with fewer artifacts within the dermis area.

\begin{table}[htbp]
    \small
    \centering
    \caption{Model comparison. MSE: Mean Squared Error; PSNR: Peak Signal-to-Noise Ratio; SSIM: Structural Similarity Index; FSIM: Feature Similarity Index; MS-SSIM: Multi-Scale Structural Similarity Index}
    \begin{tabular}{c|ccccc}
    \hline
    \textbf{Method}       & \textbf{MSE}      & \textbf{PSNR}      & \textbf{SSIM}     & \textbf{FSIM}     & \textbf{MS-SSIM}  \\ \hline
    pix2pixGAN            & 685.7    & 20.30
              & 0.4481          & 0.4026          & 0.4748          \\
    pix2pixHD             & 694.9
              & 20.45          & 0.3792          & 0.3819          & 0.4910          \\
    VSGD-Net              & 654.3
              & \textbf{20.57} & 0.4486          & 0.4025          & 0.4975          \\
    RestainNet loss       & 664.7
              & 20.35          & 0.4424          & 0.4061          & 0.4714          \\ \hline
    \textbf{In-and-Out Net} & \textbf{642.4} & 20.54          & \textbf{0.4594} & \textbf{0.4120} & \textbf{0.5081} \\ \hline
    \end{tabular}
    \label{tab:comparison}
\end{table}

\begin{table}[htbp]
    \small
    \centering
    \caption{Model comparison statistical significance test. The metrics from the listed models are conducted paired t-test with In-and-Out Net, respectively. The values here are the p-values from the tests.}
    \begin{tabular}{c|ccccc}
    \hline
    \textbf{p-values}       & \textbf{MSE}      & \textbf{PSNR}      & \textbf{SSIM}     & \textbf{FSIM}     & \textbf{MS-SSIM}  \\ \hline
    pix2pixGAN            & \textless0.01    & \textless0.01
              & \textless0.01          & \textless0.01          & \textless0.01          \\
    pix2pixHD             & \textless0.01
              & 0.0891          & \textless0.01          & \textless0.01          & \textless0.01          \\
    VSGD-Net              & 0.4984
              & 0.2343 & \textless0.01          & \textless0.01          & 0.0133          \\
    RestainNet loss       & \textless0.01
              & \textless0.01          & \textless0.01          & \textless0.01          & \textless0.01          \\ \hline
    \end{tabular}
    \label{tab:comparison test}
\end{table}

In Table~\ref{tab:comparison}, we utilized MSE and PSNR to assess the model outputs across all test group images. These metrics provide valuable insights into the statistical similarity between the output images and their corresponding ground truth images. It is worth noting that In-and-Out Net emerged with the lowest MSE, suggesting a high degree of accuracy in its predictions. Conversely, VSGD-Net showcased superior performance in PSNR, indicating its ability to preserve signal fidelity amidst noise and distortion. However, in the context of digital staining, we focus more on capturing structural details accurately. Structural similarity becomes more appropriate in determining the efficacy of the models. To this end, we employed SSIM, FSIM, and MS-SSIM to comprehensively evaluate the models' ability to preserve structural integrity. Notably, In-and-Out Net outperformed all other models in the assessment of structural similarity. This suggests its exceptional capability in capturing the intricate details crucial for digital staining applications. 

\begin{figure*}[ht]
  \centering
  \includegraphics[width=1\textwidth]{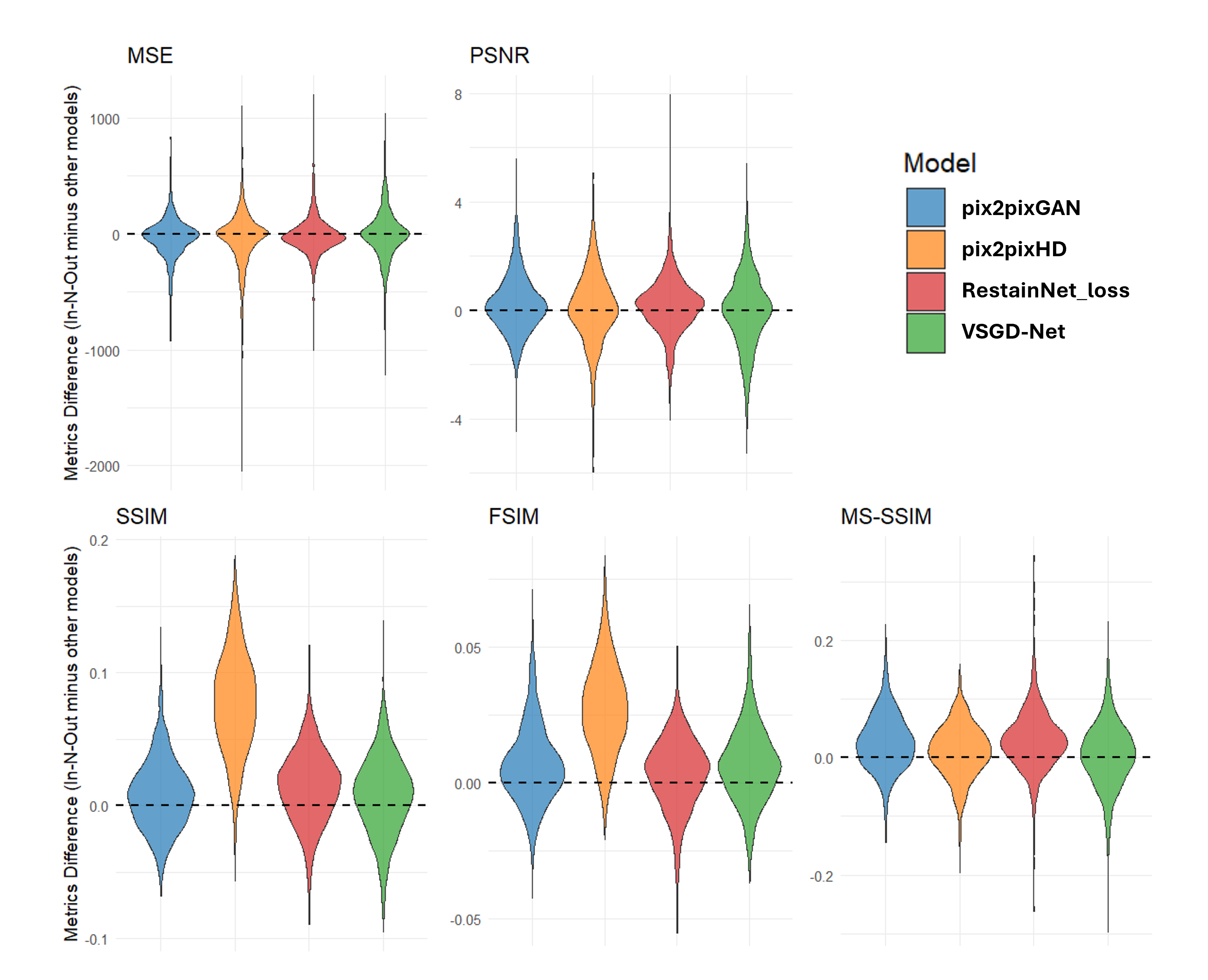}
  \caption{Violin plots comparing model metrics. We illustrate the differences between metrics obtained from the In-and-Out model and those from other models. The violin plots depict these differences.}
  \label{fig:comparison_violin}
\end{figure*}

Table~\ref{tab:comparison test} presents the paired t-test P-values comparing our In-and-Out Net with other models across several metrics. For all P-values less than $0.01$, we report them as "$\textless0.01$", demonstrating that our model's performance is statistically superior in most cases. The results show that In-and-Out Net consistently outperforms the other models, particularly in terms of SSIM, FSIM, and MS-SSIM, which are critical metrics for assessing the structural and visual fidelity of the generated images. Although the comparison with VSGD-Net reveals no significant difference in MSE and PSNR, the differences in SSIM, FSIM, and MS-SSIM are statistically significant, indicating that our model excels in capturing both the fine morphological details and color accuracy that are essential for digital staining. In Figure~\ref{fig:comparison_violin}, we visualize the differences in performance metrics between In-and-Out Net and other models using violin plots, where higher values generally indicate better performance (except for MSE). The plots highlight that In-and-Out Net achieves substantially better results, particularly for the structural metrics, where it consistently outperforms the other methods. These results strongly confirm that our approach provides a statistically significant improvement in the quality of digital H\&E stained images.

As for traditional machine learning methods, we would like to clarify that no comparison with such methods was included because, as far as we know, traditional machine learning models are not well-suited for solving the digital staining problem. These models typically excel in classification and regression tasks but lack the ability to perform pixel-wise image-to-image translations that are necessary for generating digital stained images. Furthermore, traditional machine learning methods have not been shown in the literature to successfully address the complex transformations required for digital staining. As such, no prior work has applied traditional machine learning to this task, making deep learning models, particularly Generative Adversarial Networks (GANs), the most effective and appropriate choice for this application.

\subsection{Ablation study}

\begin{table*}[h]
    \centering
    \caption{Ablation result. VOL: Variance of Laplacian; MSE: Measn Squared Error; SSIM: Structural Similarity}
    \begin{tabular}{l|cccc|lll}
    \hline
     & \textbf{In-and-Out Training} & $\textbf{D}_\textbf{out}$ & $\textbf{D}_\textbf{H}/ \textbf{D}_\textbf{E}$ & \textbf{Taking Branches} &\multicolumn{1}{c}{\textbf{VOL}}& \multicolumn{1}{c}{\textbf{MSE}} & \multicolumn{1}{c}{\textbf{SSIM}} \\ \hline
    $1$ & \XSolidBrush & \Checkmark    & \Checkmark             & \Checkmark                     & 877.1 & 654.4 & 0.4421                           \\
    $2$ & \XSolidBrush          & \XSolidBrush             & \Checkmark              & \Checkmark    & 645.3 & 660.6  & 0.4591                           \\
    $3$ & \XSolidBrush        & \Checkmark             & \textbf{\XSolidBrush}     & \Checkmark                      & \underline{44.00} &  \_\_                         & \_\_                           \\
    $4$ & \textbf{\_\_}              & \Checkmark         & \textbf{\_\_}                      & \XSolidBrush    & 608.2 & 685.7     & 0.4481      \\ \hline
    5 & \Checkmark                          & \Checkmark             & \Checkmark              & \Checkmark              & 669.1       & \textbf{642.4}               & \textbf{0.4594} \\ \hline                 
    \end{tabular}
    \label{tab:ablation}
\end{table*}

\begin{figure*}[h]
  \centering
  \includegraphics[width=0.85\textwidth]{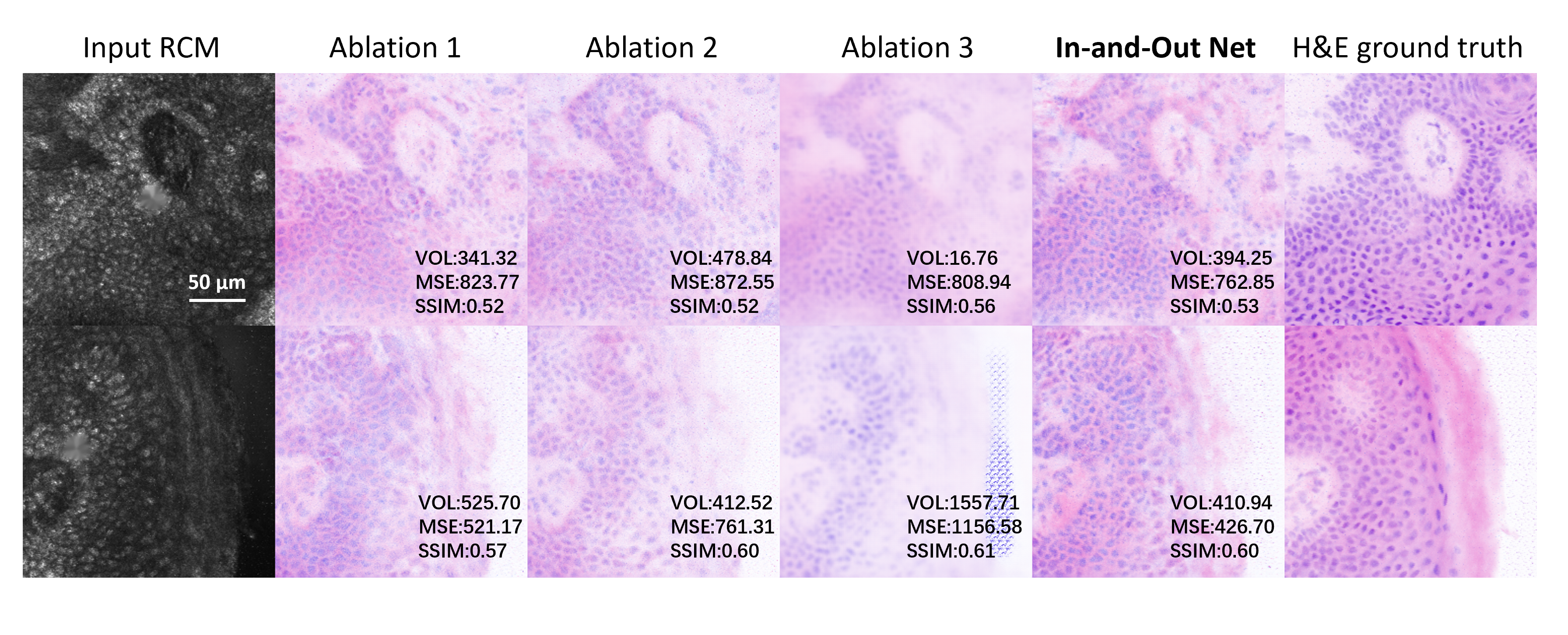}
  \caption{Examples of the ablation study. Ablation 1 removes In-and-Out training, resulting in slightly faded colors and absurd structure boundaries. Ablation 2 removes $D_{out}$, leading to faded colors. Ablation 3 removes $D_H \& D_E$, causing the image to appear blurry. The scale bar is 50 \si{\micro\metre}.}
  \label{fig:ablation}
\end{figure*}

In Table~\ref{tab:ablation}, we systematically removed each key component of the In-and-Out Net to evaluate their individual contributions. Initially, we conducted an ablation study on the In-and-Out Training, which involved training the inner and outer loops alternately. For this ablation, we trained all components simultaneously while setting $\alpha=0.5$ in Equation~\ref{eq: loss_overall}, denoted as Ablation 1. Subsequently, we performed ablations on $D_{out}$ and $D_H\&D_E$ separately, referred to as Ablation 2 and Ablation 3, respectively. Finally, we removed the branches at the beginning of the model architecture, where the RCM images were trained into two branches: H and E channel, labeled as Ablation 4. The last row in Table~\ref{tab:ablation} represents the performance of the original In-and-Out Net without any ablations. 

We employed MSE and SSIM as metrics to evaluate the results. Additionally, we introduced another metric called Variance of Laplacian (VOL) to assess the sharpness of the images. A larger VOL value indicates sharper images. The inclusion of VOL as a metric is motivated by the potential blurring effect that could occur during the ablation process, akin to out-of-focus images under a microscope. Therefore, if an image exhibits a significantly small VOL, we consider it indicative of poor sharpness and consequently disregard the model's performance. 

Table~\ref{tab:ablation} shows that the original model outperforms all ablation cases. Notably, for Ablation 3, the VOL value is significantly small, leading us to disregard its MSE and SSIM metrics due to the observed blurriness. This is reasonable because $D_H\&D_E$ were applied on H and E channel generated images during inner loop training, which focuses on learning detailed structure information. So, after ablating $D_H\&D_E$, the output images become blurry.

% \begin{table}[htbp]
%     \small
%     \centering
%     \caption{Ablation study paired t-test p-values.}
%     \begin{tabular}{c|cc}
%     \hline
%     \textbf{Ablation}       & \textbf{MSE}   & \textbf{SSIM}\\ \hline
%     1            & 685.7    & 20.30 \\
%     2            & 0.3819          & 0.4910          \\
%     3              &\_\_  & \_\_          \\
%     4       & 0.4061          & 0.4714          \\ \hline
%     \end{tabular}
%     \label{tab:ablation t-test}
% \end{table}

Figure~\ref{fig:ablation} illustrates examples of outputs from the ablation study. In Ablation 1, where In-and-Out training was omitted, the images exhibit slight blurriness, and the learned structures appear less defined. Ablation 2, which involved removing $D_{out}$, resulted in faded colors and a little ambiguous structure boundaries. In Ablation 3, where $D_H\&D_E$ were removed, the images are noticeably blurry with minimal structure details. As for Ablation 4, which mirrors the setup of pix2pixGAN, we have not included examples from this ablation in Figure~\ref{fig:ablation}.

\subsection{Model complexity comparison}
Table~\ref{tab:complexity} presents a comparison of model complexity across various methods, including FLOPs (Floating Point Operations), number of parameters, inference time, and training time. The In-and-Out Net has a higher number of parameters and FLOPs comapred to other models except pix2pixHD, which is due to its dual-branch structure, but still maintains a feasible inference time, making it applicable for clinical use. Notably, while VSGD-Net and pix2pixHD have a similar number of parameters to our model, we achieve a shorter inference time and significantly reduced training time.

A larger number of parameters generally provides greater capacity for learning from data, which is why we additionally provide a complexity comparison across the four ablation studies in Section 3.2. According to the table, Ablation 1, 2, and 3 have a similar number of parameters as the In-and-Out Net, yet Table~\ref{tab:ablation} clearly shows that In-and-Out Net achieves significantly better performance. The ablation studies highlight the effect of different components on model complexity and performance, demonstrating how each part contributes to the overall efficiency and accuracy of the model. This comparison illustrates that while In-and-Out Net has increased complexity, its performance gains are not merely due to the larger parameter count but rather the innovative architecture and training strategy employed. In addition, the inference time for a single image is sufficiently low for clinical use, meaning that despite the increased complexity of our model, it is not prohibitively high to interfere with the usability of it.
\begin{table*}[htbp]
    \small
    \centering
    \caption{Model complexity comparison across different methods. FLOP: Floating Point Operation. The table reports the FLOPs (in GFLOPs), number of parameters (in millions), inference time (in milliseconds), and training time (in hours) for each model. The In-and-Out Net shows increased complexity due to the dual-branch structure, but maintains an acceptable inference time for clinical applications. Ablation studies are included to demonstrate the contribution of different model components to the overall complexity and performance.}
    \begin{tabular}{c|cccc}
    \hline
    \textbf{Method}       & \textbf{FLOPs (Inference) (GFLOPs)}      & \textbf{Number of Parameters (Million)}      & \textbf{Inference Time (ms)}     & \textbf{Training Time (hour)}     \\ \hline
    pix2pixGAN            & 13.8    & 55.3
              & 254          & 8.01 \\
    pix2pixHD            & 54.4 & 182          & 547         & $\geq$24  \\
    VSGD-Net              & 33.7
              & 91.4 & 1551          & $\geq$48    \\
    RestainNet loss       & 13.8
              & 55.3          & 252          & 8.11 \\ \hline
    Ablation 1            & 27.7    & 111
              & 450          & 16.0 \\
    Ablation 2            & 27.7
              & 111          & 450         & 15.5  \\
    Ablation 3              & 27.7
              & 110 & 450          & 14.7    \\
    Abaltion 4            & 13.8    & 55.3
              & 254          & 8.01 \\ \hline
    \textbf{In-and-Out Net} & 27.7 & 111          & 450 & 16.2 \\ \hline
    \end{tabular}
    \label{tab:complexity}
\end{table*}

\subsection{Training approaches}
\begin{figure}[h]
  \centering
  \includegraphics[width=1\textwidth]{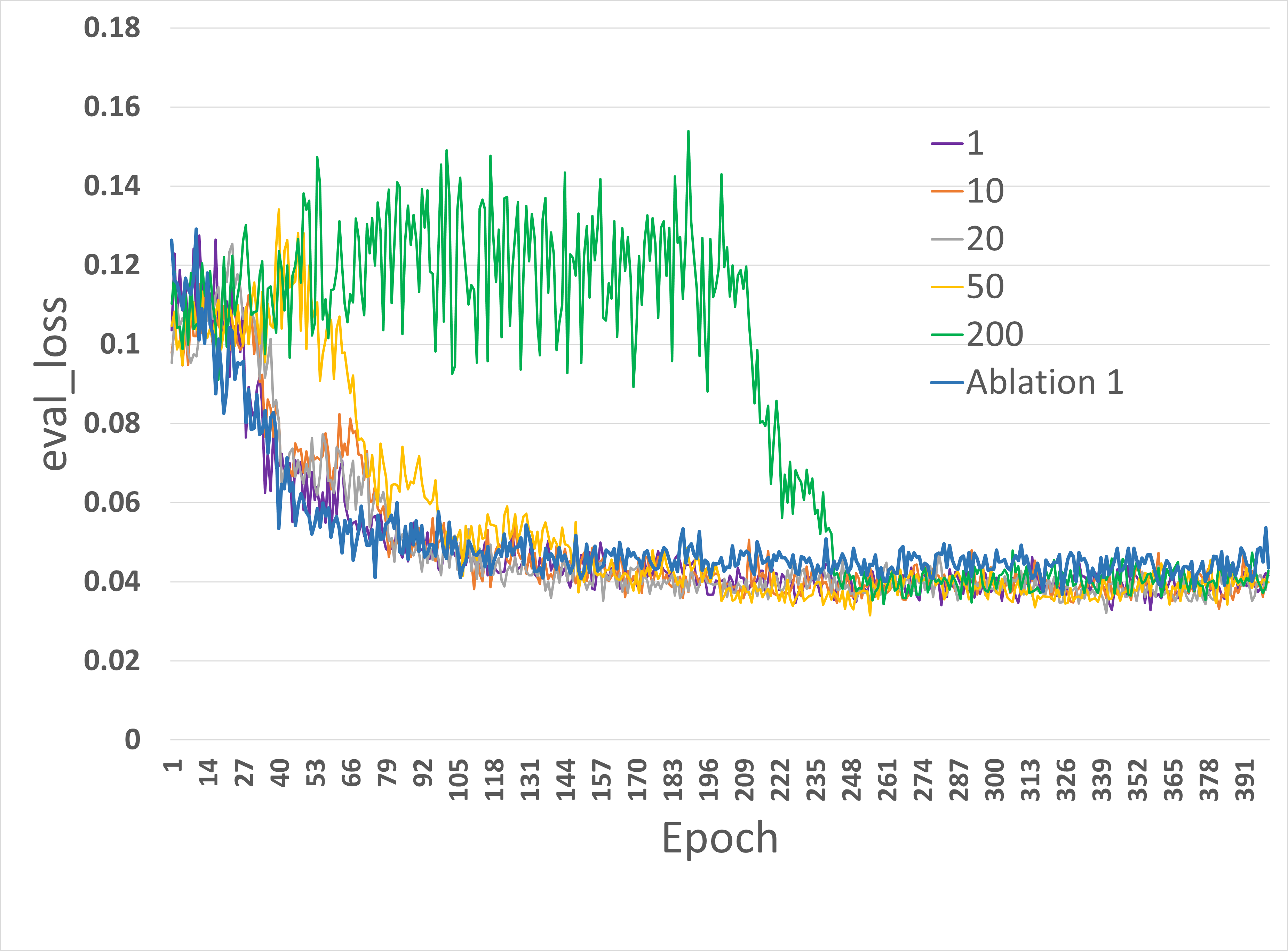}
  \caption{The change of $eval_{loss}$. Numbers in the legend indicate $n$ numbers of epochs each inner/outer loop. Step-wise descending is observed where the descending occurs during outer loops. They all converge, but Ablation 1 has a higher converge value.}
  \label{fig:training}
\end{figure}

Figure~\ref{fig:training} illustrates the impact of different training approaches on $eval\_loss$, as defined in Equation~\ref{eq: eval_loss}. Our training methodology involves alternating between $n$ epochs of inner loop training followed by $n$ epochs of outer loop training, and repeating this cycle until a total of $400$ epochs have been completed. In Figure~\ref{fig:training}, the values of $n$ are indicated in the legend. Additionally, we include the results of Ablation 1 in the figure for comparison, as it was trained without the In-and-Out training approach.

The plot reveals a distinct step pattern in the decrease of $eval\_loss$ when $n$ is relatively large, such as $200$ and $50$. Notably, training the outer loop leads to a significant decrease in $eval\_loss$. The reason for this observation is that during inner loop training, color and channel combination information is not fully utilized. Nevertheless, inner loop training remains essential for stabilizing structural information. Conversely, Ablation 3 only trains the outer loop, which demonstrates the importance of incorporating both inner and outer loop training for optimal results.

When $n$ is small, the stepwise reduction in $eval\_loss$ is less obvious, but convergence is still achieved across all training approaches. However, comparing the curve of Ablation 1 to the In-and-Out training approaches, it becomes evident that while Ablation 1 can converge, it remains a relatively higher $eval\_loss$ value, indicating the effectiveness of the In-and-Out training approach in minimizing $eval\_loss$.
 
\subsection{Discussion}
In-and-Out Net effectively transforms RCM images into H\&E-stained images with remarkable color fidelity and accurate structural representation. Unlike other models, it is purposefully designed for digital staining tasks. In this study, we produced H\&E ground truth images comprising two distinct fluorescence channels, enabling the separate training of both channels. However, even with real H\&E-stained images as ground truth, we can utilize color decomposition to isolate the H and E channel images for employment with In-and-Out Net. The good performance of In-and-Out Net comes from its utilization of the prior knowledge that H\&E-stained images consist of two fixed colors. Moreover, it employs a dual-stage training approach, comprising coarse training (inner loop) and fine training (outer loop), to effectively converge towards an optimal local minimum for the loss function.

A limitation of our study is the inadequate quality of RCM images. As shown in Table~\ref{tab:comparison}, even the highest SSIM attained is merely 0.5081. This hampers our ability to morphologically compare specific structures accurately, primarily due to the poor quality of RCM images utilized. In most studies, investigators would employ commercial RCM systems such as the VivaScope 1500 System (Caliber I.D., Rochester, NY) for imaging, which offer superior Signal-to-Noise Ratio (SNR), enhanced contrast, superior resolution, and more detailed structural information compared to our dataset. One potential explanation could be the disparity in laser wavelengths, as we used a 488 nm laser for RCM imaging, while the Caliber system utilizes an 830 nm laser. Additionally, the system we employed is not specifically designed for RCM, leading to some optical limitations for reflectance mode. Owing to these system limitations, acquiring high-quality RCM data is challenging, which results in most images captured from dermal/epidermal junctions. So, our analysis lacks thorough examination within different tissue categories. Nonetheless, these limitations can be addressed with improved RCM inputs, and they do not undermine the efficacy of our In-and-Out Net methodology. While we've shown that creating ground truth with two fluorescence images is feasible, it's essential for clinical experts to validate this approach in the future. This validation will help determine if there are significant differences in information between the pseudo ground truth and actual H\&E staining images.

This study pioneers the application of the In-and-Out Net model, demonstrating its effectiveness in digitally converting RCM images to H\&E staining counterparts, utilizing data from skin BCC cases. Future investigations may explore its applicability to other single color image types. Additionally, we would also test it on other tumors (such as squamous cell carcinoma, Merkel cell carcinoma) and skin conditions like psoriasis. Furthermore, there's potential to assess its versatility across different histological staining techniques in subsequent studies.

\section{Conclusion}
In this study, we introduce a novel digital staining network called In-and-Out Net. We utilize this network to digitally stain RCM images to H\&E-stained images and compare its performance with other networks. Additionally, we conduct an ablation study to validate the functionality of each component within our network. Furthermore, we compare various training approaches. Through the utilization of In-and-Out Net, we achieve successful staining of RCM images to H\&E with excellent color and structural matching, outperforming other models. Through the ablation study, we confirm the importance of In-and-Out training and validate the functions of the inner loop and outer loop by ablating three discriminators. Moreover, we investigate the effects of different training approaches. Despite the limitations imposed by the quality of RCM images, we are the first to digitally stain aluminum-processed RCM images to H\&E images. We anticipate that In-and-Out Net can be widely applied to various digital staining tasks.

\section*{Acknowledgements}
Research reported in this publication was supported by the National Cancer Institute of the National Institutes of Health under Award Number R01CA273734. The content is solely the responsibility of the authors and does not necessarily represent the official views of the National Institutes of Health. We plan to share the code used in this study on GitHub for the benefit of the research community. The code will be organized and uploaded to GitHub following the publication of this paper.

% \bibliographystyle{elsarticle-num-names}
\bibliographystyle{plainnat}
\bibliography{library}



%% For example:

%% \bibitem[Aladro et al.(2015)]{Aladro15} Aladro, R., Martín, S., Riquelme, D., et al. 2015, \aas, 579, A101

%%\end{thebibliography}

\end{document}